# Automated Pavement Cracks Detection and Classification Using Deep Learning

Selvia Nafaa
*Computers and Systems Department*
*Minia University*
Minia, Egypt

Karim Ashour
*Computers and Systems Department*
*Minia University*
Minia, Egypt

Rana Mohamed
*Computers and Systems Department*
*Minia University*
Minia, Egypt

Hafsa Essam
*Computers and Systems Department*
*Minia University*
Minia, Egypt

Doaa Emad
*Computers and Systems Department*
*Minia University*
Minia, Egypt

Mohammed Elhenawy
*CARRS-Q*
*Queensland University of Technology*
Brisbane, Australia
mohammed.elhenawy@qut.edu.au

Huthaifa I. Ashqar
*Civil Engineering Department*
*Arab American University*
Jenin, Palestine
huthaifa.ashqar@aaup.edu

Abdallah A. Hassan
*Computers and Systems Department*
*Minia University*
Minia, Egypt
abdallah@mu.edu.eg

Taqwa I. Alhadidi
*Civil Engineering Department*
*Al-Ahliyya Amman University*
Amman, Jordan
t.alhadidi@ammanu.edu.jo

*Abstract*—Monitoring asset conditions is a crucial factor in building efficient transportation asset management. Because of substantial advances in image processing, traditional manual classification has been largely replaced by semi-automatic/automatic techniques. As a result, automated asset detection and classification techniques are required. This paper proposes a methodology to detect and classify roadway pavement cracks using the well-known You Only Look Once (YOLO) version five (YOLOv5) and version 8 (YOLOv8) algorithms. Experimental results indicated that the precision of pavement crack detection reaches up to 67.3% under different illumination conditions and image sizes. The findings of this study can assist highway agencies in accurately detecting and classifying asset conditions under different illumination conditions. This will reduce the cost and time that are associated with manual inspection, which can greatly reduce the cost of highway asset maintenance.

## I. Introduction

In the area of infrastructure maintenance, you must check for, examine and inspect faults in pavement surfaces in the interests of the safety and longevity of roads and highways. This has been a manual process, traditionally, which means that not only does it take up free time, but is also subject to human error. The implementation of computer vision has greatly revolutionised this field, providing us with a more efficient, accurate and automated method of crack detection [1].

Computer vision is a subdivision of artificial intelligence which aims to teach computers how to interpret and understand what they see. Computer vision systems, using digital images and deep learning models, can detect patterns and anomalies in ways that far surpass the accuracy and efficiency of human inspectors. These systems, with respect to pavement care, analyse images to find cracks; they then differentiate them by size and severity. According to [2] and [3], this information is important for maintenance teams because it allows them to assign repair work in the most logical order and to get the most for their efforts.

In an urban road network managing system for assets, achieving success requires a powerful and continued emergence of pavement and roadside infrastructure data from the field [4]. An ingredient of asset management in roads is the process of asset inventory and field data collection. Asset inspection was made on three types of assets: bridges, pavement cracks, and signs [5]. However, the inventory in general, also includes location, condition, and detailed information on each object. The old-fashioned way of collecting data was very labour-intensive, potentially dangerous, and took a long time for various things that happened during manual approaches using handheld instruments [6]. To overcome these difficulties, researchers need to adopt new cutting-edge technics and methods for data collection. Specialised vehicles equipped with GPS (spatial data), photograph and video cameras (inventory and distress data), and many sensors such as laser scanners, profilers, and accelerometers can be utilised for automated data collecting [7], [8].

Deep learning used in asset management and detection systems is successful nowadays due to advances in artificial intelligence and computer hardware [9], [10], [11]. Several different techniques are used in pavement crack detection, including Convolutional neural networks (CNN) [12], [13], [14]. Two-stage CNN [15], faster R-CNN [16], a Network CNN [13] and others of its kind exist to aid in pavement crack detection. The faster processing involves less fine-tuning. Moreover, signs were detected and classified in several works [5], [17], [18]. These algorithms have produced good results only in a perfect environment, while, in fact, they were all similar to the model described above. Most of it, however, was disappointed with their performance when used outside of a controlled setting. The performance of the interference model of environmental elements was not considered in all these models. This paper uses modified YOLOv5 and YOLOv8 [9] in order to improve the detection accuracy, detection speed, and robustness of detection methods when the brightness changes [19].

This process involves the application of artificial neural networks, and deep learning has made meaningful progress in computer vision to detect cracks in paths [20]. Convolutional neural networks (CNNs) have displayed tremendous impact in automatically identifying cracks on streets [21]. A number of studies have researched various classification algorithms used in computer vision systems designed to detect pavement splits, as shown in multiple reports [22]. A comprehensive account of how machine vision technology works in identifying pavement cracks is presented in the literature. This involves different stages: the capture of cracked images, preprocessing, segmentation, and application of recognition technology [23]. Deep learning, such as deep convolutional neural networks (CNNs), has recently become a popular means of automated pavement distress detection because of its dramatically enhanced efficacy and accuracy[24]. Deep learning has become widespread in modern times, and with it, the Feature Pyramid and Hierarchical Boosting Network (FPHBN)[25]. This means contemporary methods utilise image analysis and computer vision tools as a means to automate the identification of fractures in asphalt[26] . Finally, work has also shifted to novel vision-based methods for detecting fractures in pavements. A number of researchers have improved the use of better YOLOv5s to search out diseases in asphalt pavement [27].

This paper designs an effective pavement crack detection and classification based on the theoretical knowledge of deep learning to improve the detection accuracy, detection speed, and robustness of detection methods under different illumination conditions.

## II. METHODOLOGY

This study aims to propose a novel framework that can classify the cracks in road pavement. For this reason, we applied YOLOv5 and YOLOv8 to detect and classify pavement cracks using the RDD2022 open-source dataset. RDD2022 consists of 47,420 road photos that were gathered in China, Japan, the Czech Republic, India, and the United States in order to suggest techniques for autonomous road damage detection in various nations. Smartphones were used to record the dataset, and Google Street View images were taken using high-resolution cameras from cars, motorcycles, and drones. The distribution of the pavement distresses in the original data is shown in Fig. 1.

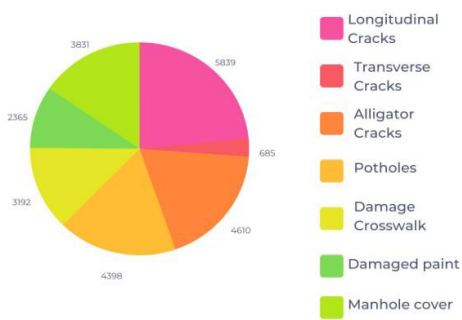

Fig. 1. Dataset classes and number of objects in each class.

However, because we have a limited amount of memory space and big data training model resources, in the pavement crack detection and classification, we only used 8,535 images with a 600 × 600 pixels resolution images from the RDD data that was collected from Japan. We used the seven classes shown in Fig. 1, with about 70% of the dataset as training, 20% as validation, and 10% as testing. The specific parameters of the workstation for the experiment include GPU of NVIDIA GeForce RTX 3070 Ti, CPU of AMD Ryzen 5 56000X 6-core processor, and RAM of 16.0 GB.

In facilitating object detection, several image processing techniques were implemented in this work. Specifically, cutting the higher edge and blackout background are used in pavement crack detection, while different illumination settings were used in signs detection. In the pavement crack detection, we noticed that most of the crack location is located at the bottom of the image. As such, we tested different cutting-edge sizes to improve the accuracy of the pre-trained model accuracy. This cutting edge helps the model perform better and minimises computational time. After cutting the edge, we blackout the background elements to focus on pavement defects only. This was done by separating elements outside of the pavement's colour range (given as lower range = [127, 36, 33] and upper range = [179, 255, 255]). By superimposing bounding boxes on photos and ensuring that the cracks remained distinct from other image components, we validated the efficacy of our method. Fig. 2 demonstrates that our model consistently displayed proficiency in defect identification. We also cut the top area from the images of the original not-segmented dataset as the pavement that contains the defects located at the lower part of the image. Most images in the dataset have a size of 600x600 pixels. We tested two versions of cutting, including 600x420 pixels and 600x330 pixels. We found that these ranges contain the pavement in most images. Both have very close results and almost the same accuracy, the different crops are shown in Fig. 3.

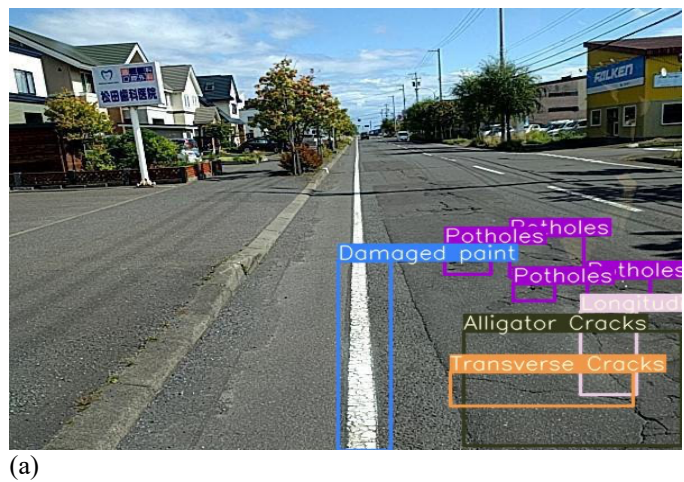

(a)

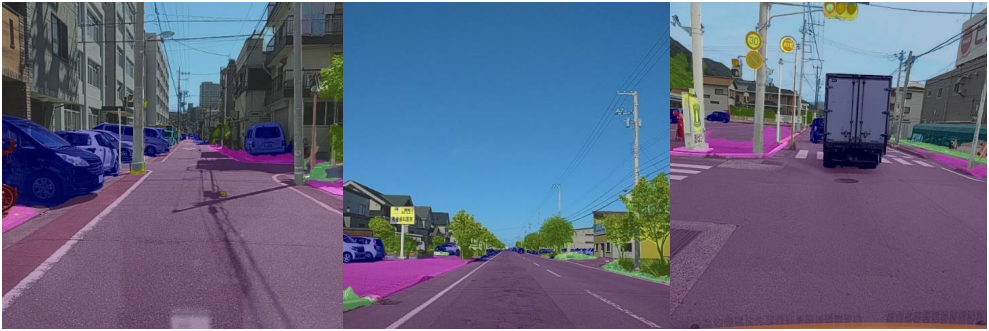

(b)

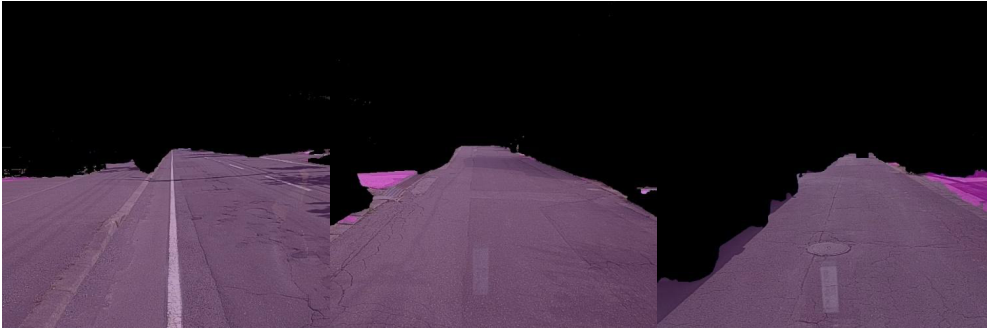

(c)

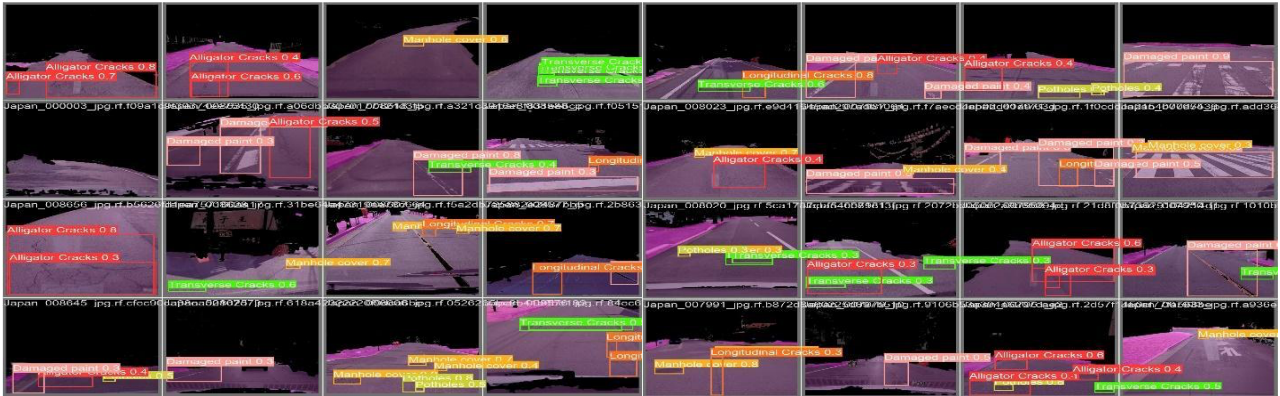

(d)

Fig. 2. Pavement Crack Blackout and detection steps. (a) original pavement images, (b) pre-trained model without blackout results, (c) Blackout results, (d) validation of pavement crack detection and classification.

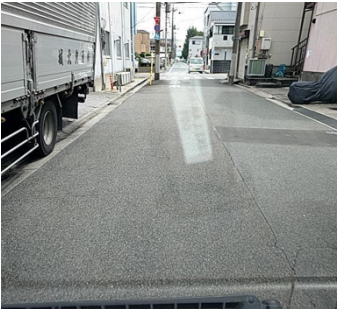
(a)

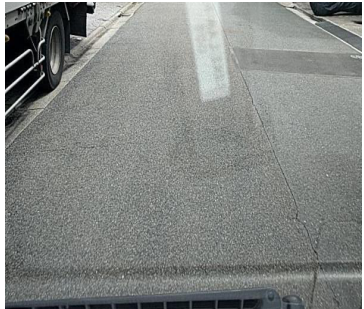
(b)

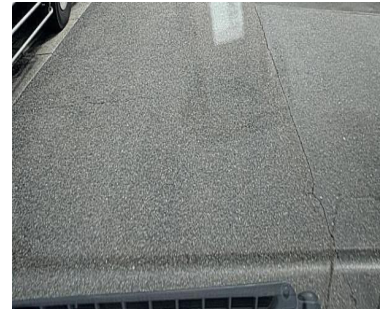
(c)

Fig. 3. Pavement cutting edge. (a) original pavement images, (b) 600× 420 pixels, (c) 600× 330 pixels.

### III. ANALYSIS AND RESULTS

In this section, the two developed models are presented using evaluation metrics and visual results. In order to demonstrate the effectiveness of the YOLOv5 and YOLOv8 in pavement crack detection, a balanced dataset for testing was used, and the model evaluation metrics for each model is shown in Table 1. This is what we will consider the baseline performance.

Table 1. Evaluation metrics for baseline model.

| Model | MAP | Precision | Recall | F1-Score |
|---|---|---|---|---|
| **YOLOv5** | 61.6% | 64.6% | 60.1% | 62.3 |
| **YOLOv8** | 62.2% | 65.0% | 61.0% | 62.9 |

Several enhancements, such as blackout, weighted segmentation, and different cropping sizes, were implemented to improve the model performance. Table 2 shows the models' performance after various improvements using various metrics.

Table 2 shows that different classification algorithms have different performances at different improvement techniques. In the following paragraphs, we will analyse the results of Table 2 in comparison with Table 1 (i.e. baseline):

**Merging Classes:** When compared to the baseline, this method decreased the MAP, Recall, and F1-Score for both models. For YOLOv5, precision declined marginally, but for YOLOv8, it stayed almost the same.

**Blackout:** All metrics for YOLOv5 showed a modest decline. The MAP and precision of YOLOv8 were better than the baseline, but a minor decline in Recall and F1-Score. This implies that blackout as a strategy might help YOLOv8 with precision and MAP.

**Blackout Merging:** YOLOv5 and YOLOv8 performed worse than the baseline in all metrics except the precision of YOLOv8 when the blackout and merging class strategies were combined. The effect is particularly noticeable in recall and MAP.

Table 2. Models' performance with several improvement techniques.

| Technique | Model | MAP | Precision | Recall | F1-Score |
|---|---|---|---|---|---|
| Merging Classes | YOLOv5 | 58.3% | 63.0% | 55.3% | 58.9 |
| | YOLOv8 | 60.0% | 64.4% | 56.7% | 60.3 |
| Blackout | YOLOv5 | 60.5% | 63.7% | 58.9% | 61.2 |
| | YOLOv8 | 62.4% | 66.1% | 58.8% | 62.2 |
| Blackout Merging | YOLOv5 | 57.2% | 63.0% | 54.2% | 58.3 |
| | YOLOv8 | 58.8% | 67.3% | 54.1% | 59.9 |
| First Cropping Size | YOLOv5 | 61.5% | 60.7% | 63.5% | 62.1 |
| | YOLOv8 | 63.4% | 66.9% | 58.7% | 62.5 |
| Second Cropping Size | YOLOv5 | 60.9% | 60.9% | 59.3% | 60.1 |
| | YOLOv8 | 63.0% | 67.3% | 58.7% | 62.7 |

**First Cropping Size:** When compared to the baseline, this method exhibits a drop in MAP precision and F1-Score for YOLOv5 and an increase in precision. For YOLOv8, precision and MAP have increased, while recall and MAP, on the other hand, have decreased.

**Second Cropping Size:** Compared to the baseline, there is a drop in all evaluation metrics for YOLOv5. In the case of YOLOv8, precision and MAP have increased, while recall and MAP, on the other hand, have decreased.

Results show that sophisticated deep learning algorithms, in particular YOLOv5 and YOLOv8, have the potential to be used for the identification of pavement cracks. Based on the findings that were gathered, we found that the implementation of YOLOv5 and YOLOv8 in our research has demonstrated an improvement in the accuracy of pavement crack detection. This demonstrates the efficacy of the YOLO algorithms. These findings shed light on the resiliency of these algorithms, which is especially important when one considers the difficulties presented by the shifting illumination and picture perspectives that occur in real-world situations.

IV. CONCLUSION

A roadway asset identification method based on a deep convolutional neural network fusion model was implemented in this study. The model validation strategy was carried out by testing the performance of several models for detecting roadway signs and pavement cracks, effectively improving its asset detection accuracy. In highway asset management, our findings suggest a considerable improvement compared to typical methods. The use of more sophisticated algorithms in this research hints at a strategy for the maintenance of roadways that is more precise, efficient, and cost-effective. This strategy is essential for ensuring the longevity of infrastructure and protecting the public's safety.

The successful implementation of these algorithms in our study implies their potential integration into real-time roadway monitoring systems. This integration could lead to transformative changes in roadway asset management, enabling more proactive and efficient maintenance practices. The study paves the way for further exploration in the domain of AI applications in transportation. Future research might focus on enhancing the algorithms' performance in diverse environmental conditions, integrating these techniques with other technological advancements, and exploring their applications in broader aspects of transportation infrastructure management.